\pdfoutput=1
\documentclass[sigconf]{acmart} 

\usepackage{booktabs} 
\usepackage{subfig}
\usepackage{xcolor}
\usepackage{color, colortbl}
\usepackage{float}

\setcopyright{none}

\newcommand{\hk}[1]{{\color{black} #1}} 

\renewcommand\footnotetextcopyrightpermission[1]{} 
\begin{document}
\title{A Neural Network Approach to Missing Marker Reconstruction in Human Motion Capture}

\author{Taras Kucherenko}
\orcid{0000-0001-9838-8848}
\affiliation{%
  \department{Robotics, Perception and Learning}
  \institution{KTH Royal Institute of Technology}
  \country{Sweden}}
\email{tarask@kth.se}

\author{Jonas Beskow}
\orcid{0000-0003-1399-6604}
\affiliation{%
  \department{Speech, Music and Hearing}
  \institution{KTH Royal Institute of Technology}
  \country{Sweden}}
\email{beskow@kth.se}

\author{Hedvig Kjellstr{\"o}m}
\orcid{0000-0002-5750-9655}
\affiliation{%
  \department{Robotics, Perception and Learning}
  \institution{KTH Royal Institute of Technology}
  \country{Sweden}}
\email{hedvig@kth.se}

\renewcommand{\shortauthors}{T. Kucherenko et al.}

\begin{abstract}
Optical motion capture systems have become a widely used technology in various fields, such as augmented reality, robotics, movie production, etc. Such systems use a large number of cameras to triangulate the position of optical markers.
The marker positions are estimated with high accuracy. However, especially when tracking articulated bodies, a fraction of the markers in each timestep is missing from the reconstruction. 

In this paper, we propose to use a neural network approach to learn how human motion is temporally and spatially correlated, and reconstruct missing markers positions through this model.  
We experiment with two
different models, one LSTM-based and one time-window-based. Both methods produce state-of-the-art results, while working online, as opposed to most of the alternative methods, which require the complete sequence to be known. The implementation is publicly available at https://github.com/Svito-zar/NN-for-Missing-Marker-Reconstruction.
\end{abstract}

%

\ccsdesc[500]{Computing methodologies~Machine learning}
\ccsdesc[500]{Computing methodologies~Motion processing}
\ccsdesc[300]{Computing methodologies~Neural networks}
\ccsdesc[300]{Computing methodologies~Motion capture}

\keywords{Motion capture, Missing markers, Neural Networks, Deep Learning}

\maketitle


\pdfoutput=1
\section{Introduction}
Often a digital representation of human motion is needed. This representation is useful in a wide range of scenarios: mapping an actor performance to a virtual avatar (in movie productions or in the game industry); predicting or classifying a motion (in robotics)
; trying clothes in a digital mirror; etc.

A common way to obtain this digital representation is marker-based optical motion capture (mocap) systems. Such systems use a large number of cameras to triangulate the position of optical markers.These are then used to reconstruct the motion of the objects to which the markers are attached.

All 
motion capture systems 
suffer to a higher or lower degree from missing marker detections, due to occlusion problems (less
than two cameras see the marker) or marker detection failures. 

\begin{figure}[t]
\centering 
\includegraphics[width=0.99\linewidth]{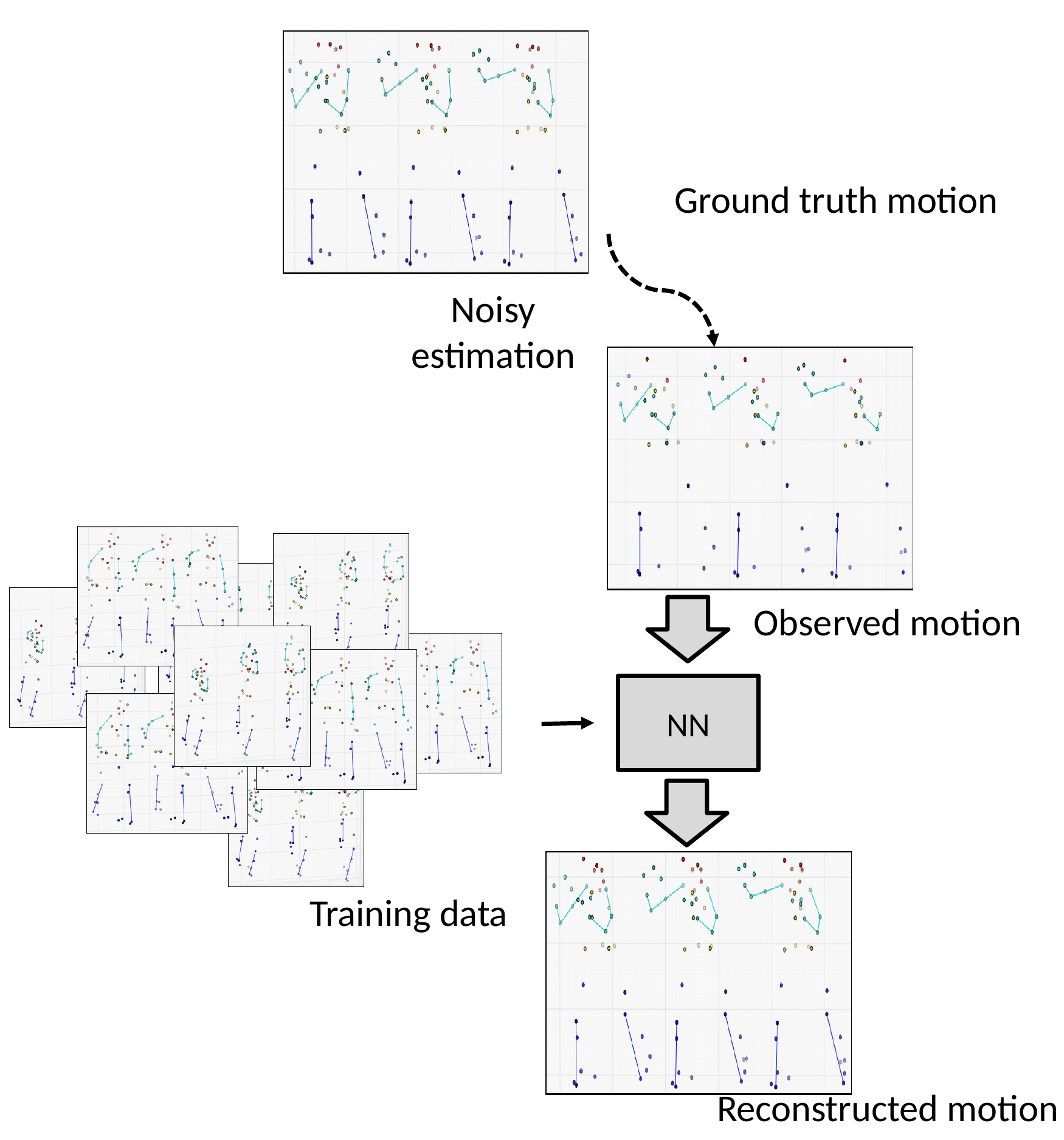} %
\caption{Illustration of our method for missing marker reconstruction. Due to errors in the capturing process, some markers are not captured. The proposed method exploits spatial and temporal correlations to reconstruct the pose of the missing markers.}
\label{fig:intro}
\end{figure}

In this paper, we propose a method for 
reconstruction of missing markers to create a more complete pose estimate (see Figure~\ref{fig:intro}). The method exploits knowledge about spatial and temporal correlation in human motion, learned from data examples to remove position noise and fill in missing parts of the pose estimate (Section~\ref{secMeth}).


A number of methods have been proposed within the Graphics community to address the problem of 
missing marker reconstruction. The traditional approach  \cite{burke2016estimating,peng2015hierarchical} is interpolation within the current sequence. Wang et al proposed a method which exploits motion examples to learn typical correlations \cite{wang2016human}. The novelty of our method with respect to theirs is that while they learn linear dependencies, we employ a Neural Network (NN) methodology which enables modeling of more complicated spatial and temporal correlations in sequences of a human pose.

In Section ~\ref{secExp} we demonstrate the effectiveness of our network, showing that our method outperforms the state of the art in missing marker reconstruction in various conditions.

Finally, we discuss our results in Section~\ref{Sec:Disc}. 


\pdfoutput=1
\section{Related work}
\label{sec:relWork}
The task of modeling human motion from mocap data has been studied quite extensively in the past. We here give a short review of the works most related to ours.

\subsection{Missing Marker Reconstruction} 
It is possible to do 3D pose estimation even with affordable sensors such as Kinect. However, all motion capture systems suffer to some degree from missing data. 
This has created a need for methods for \textit{missing marker reconstruction}.

The missing marker problem has been traditionally formulated as a matrix completion task. Peng et al.~\cite{peng2015hierarchical} solve it by non-negative matrix factorization, using the hierarchy of the body to break the motion into blocks. Wang et al.~\cite{wang2016human} follow the idea of decomposing the motion and do dictionary learning for each body part. They train their system separately for each type of motion. Burke and Lasenby \cite{burke2016estimating} apply PCA first and do Kalman smoothing afterward, in the lower dimensional space. Gloersen and Federolf \cite{gloersen2016predicting} used weighted PCA to reconstruct markers. Taylor et al. \cite{taylor2007modeling} applied Condition Restricted Boltzmann Machine based on the binary variables. Both Taylor \cite{taylor2007modeling} and Gloersen\cite{gloersen2016predicting} were limited to cyclic motions, such as walking and running. All those methods are based on linear algebra. They make strong assumptions about the data: each marker is often assumed to be present at least at one time-step in the sequence. 
Moreover, due to the linear models, they often struggle to reconstruct irregular and complex motion. 

The limitations discussed above motivate the neural network approach to the missing marker problem. Mall et al. \cite{mall2017deep} successfully applied deep neural network based on the bidirectional LSTM to denoise human motion and recover missing markers. Our approach is similar to theirs, but we are using a simpler network, which requires less data and computational resources. We also experiment with two different ways to handle the sequential character of the problem, while they just choose one approach.  

\subsection{Denoising}
Another related task which can be tackled with our networks is removing additive noise from the marker data. Recently Holden \cite{holden2018robust} used a similar approach to this problem. He employed a neural network that took noisy markers as an input and returned clean body position as an output. The main difference is that our network is also capable of reconstructing missing values and takes sequential information into account.

\subsection{Prediction}
A highly related problem is to predict human motion some time into the future. 

State-of-the-art methods try to ensure continuity either by using Recurrent Neural Networks (RNNs) \cite{fragkiadaki2015recurrent,jain2016structural} or by feeding many time-frames at the same time \cite{butepage2017deep}. While our focus is not on prediction, our networks architectures are inspired by those methods. 

Since our application is not a prediction, our architecture is slightly different. 

Another related paper is the work of B\"utepage et al.~\cite{butepage2017deep}, who use a sliding window and a Fully Connected Neural Network (FCNN) to do motion prediction and classification. Again, since our problem is different, we modify their network, 
using a much shorter window length, fewer layers, and no bottleneck.


\pdfoutput=1
\section{Method overview}
\label{secMeth}

In the following section, we give a mathematical problem formulation and an overview of the proposed approach.

\subsection{Missing Markers}
\label{subsec_Noise}

Missing markers in real life correspond to the failure of a sensor in the motion capture system.

In our experiments, we use mocap data without missing markers. 
Missing markers are emulated by nullifying some marker positions in each frame. This process can be mathematically formulated as a multiplication of the mocap frame $\mathbf{x}_t$ by a binary matrix $M_t$: 
\begin{equation}
\label{eq:missing}
\mathbf{\hat{x}}_t = C(\mathbf{x}_t) = M_t  \mathbf{x}_t,
\end{equation}
where $M_t \in [0,1] ^{3n x 3n}$, such that that all 3 coordinates of any marker are either missing or present at the same time. 

Every marker is missing over a few time-frames. The percentage of missing values is usually referred to as the \textit{missing rate}.


\subsection{Missing Marker Reconstruction as Function Approximation}

Missing marker reconstruction is defined in the following way: Given a human motion sequence $\mathbf{\hat{x}}$ corrupted by missing markers, the goal is to reconstruct the true pose $\mathbf{x}_t$ for every frame $t$.

We approach missing markers reconstruction as a function approximation problem: The goal is to learn a reconstruction function $R$ that approximates the inverse of the corruption function $C$ in Eq.~(\ref{eq:missing}). This function would map the sequence of corrupted poses to an approximation of the true poses:
\begin{equation}
\mathbf{x} = R(\mathbf{\hat{x}}) \approx C^{-1}(\mathbf{\hat{x}})
\end{equation}
The mapping $C$ is under-determined, so it is not invertible. However, it can be approximated by learning
spatial and temporal correlations in human motion in general, from a set of other pose sequences. 

We propose to use a Neural Network (NN) approach to learn $R$, well known for being a powerful tool for function approximation \cite{hornik1989multilayer}.
We employ two different types of neural network models, which are described in the following sections. Both of them are using a principle of Denoising Autoencoder \cite{vincent2008extracting}: during the training Gaussian additive noise was injected into the input:
\begin{equation}
\label{eq:missing}
\mathbf{\hat{x}}_t = \hat{C}(\mathbf{x}_t) = M_t ( \mathbf{x}_t + \mathcal{N}(0, \sigma(X)*\alpha)),
\end{equation}
where $\sigma(X)$ is a standard deviation in the training dataset and $\alpha$ is a coefficient of proportionality, which we call \textit{the noise parameter}. It was experimentally set to the value of 0.3.

Denoising is commonly used to regularize encoder-decoder NN. Experiments proved it to be beneficial in our application as well.

The network was learning to remove noise, at the same time as reconstructing missing values. During the testing, no noise was injected. Our two methods are compared to each other and to the state of the art in missing marker reconstruction in Section~\ref{secExp}.

\section{Neural Network Architectures}

In this section, the two versions of the method are explained. 

\begin{figure}
\centering
\subfloat[LSTM-based]{~~~\includegraphics[width=0.27\linewidth]{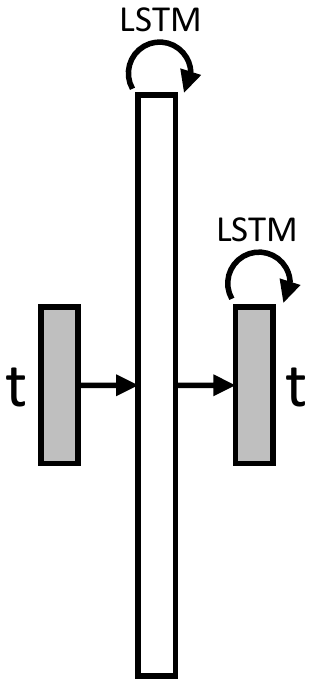}~~~}~~
\subfloat[Window-based]{~~~\includegraphics[width=0.35\linewidth]{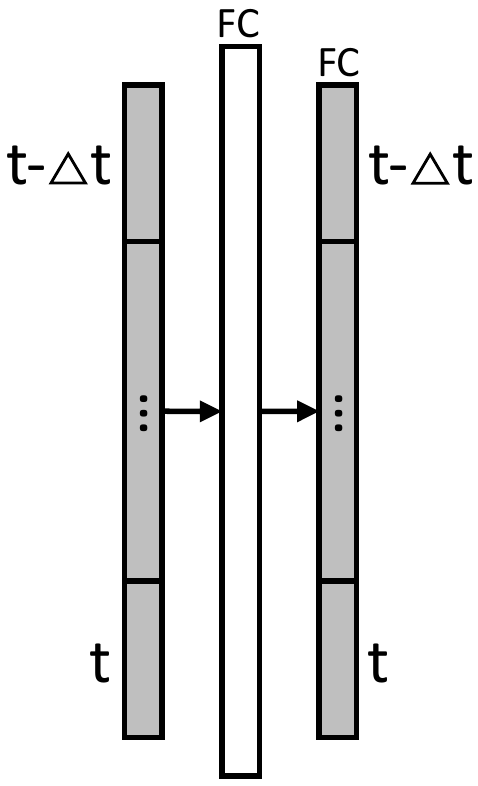}~~~}
\vspace{-2mm}
\caption{Illustration of the two architecture types. (a) LSTM-based architecture (Section~\ref{subsec:LSTM}). (b) Window-based architecture (Section~\ref{subsec:Window}).}
\label{fig:models}
\end{figure}

\subsection{LSTM-Based Neural Network Architecture}
\label{subsec:LSTM}

Long-Short Term Memory (LSTM) \cite{lstm1997} is a special type of Recurrent Neural Network (RNN). It was designed as a solution to the vanishing gradient problem \cite{hochreiter1998vanishing} and has become a default choice for many problems that involve sequence-to-sequence mapping \cite{sutskever2014sequence, donahue2015long, shi2017end}.

Our network is based on LSTM and illustrated in Figure~\ref{fig:models}a. The input layer is a corrupted pose $\mathbf{\hat{x}}_t$, and the output LSTM layer is the corresponding true pose $\mathbf{x}_t$. 


%
\subsection{Window-based Neural Network architecture}
\label{subsec:Window}
An alternative approach is to use a range of previous time-steps explicitly, and to train a regular Fully Connected Neural Network (FCNN) with the current pose along with a short history, i.e., a window of poses over time $(t - \Delta t):t$. 


This network is illustrated in Figure~\ref{fig:models}b. The input layer is a window of concatenated corrupted poses $[\mathbf{\hat{x}}^T_{t-\Delta t}, ..., \mathbf{\hat{x}}^T_t]^T$. The output layer is the corresponding window of true poses $[\mathbf{x}^T_{t-\Delta t}, ..., \mathbf{x}^T_t]^T$. In between, there are a few hidden fully connected layers. 

This structure is inspired by the sliding time window-based method of B\"utepage et al.~\cite{butepage2017deep}, but is adapted to pose reconstruction. For example, there is no bottleneck middle layer and fewer layers in general, to create a tighter coupling between the corrupted and real pose, rather than learning a high-level \hk{and} holistic mapping of a pose. We also use window length T=10, instead of 100, based on the performance on the validation dataset.




\section{Dataset}
\label{sec:Data}

We evaluate our method on the popular benchmark CMU Mocap dataset \cite{CMUdata}. This database contains
2235 mocap sequences of 144 different subjects. We use the recordings of 25 subjects, sampled at the rate of 120 Hz, covering a wide range of activities, such as boxing, dancing, acrobatics and running.

\begin{figure}[t]

\centering
\includegraphics[width=0.45\linewidth]{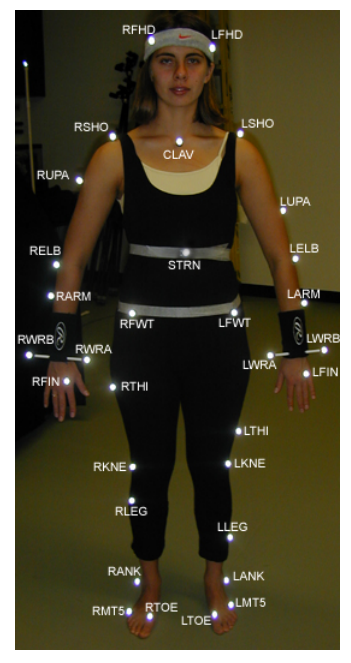}
\vspace{-2mm}
\caption[Marker placement]{Marker placement in the CMU Mocap dataset (mocap.cs.cmu.edu).}
\label{fig:markers}

\end{figure}
\vspace{-2mm}

\subsection{Preprocessing}
We start preprocessing by transforming every mocap sequence into the hips-center coordinate system. First joint angles from the BVH file are transformed into the 3D coordinates of the joints. The 3D coordinates are translated to the center of the hips by subtracting hip coordinates from each marker.
We then normalize the data into the range [-1,1] by subtracting the mean pose over the whole dataset and \hk{then dividing all values by the} absolute maximal value in the dataset.

\subsection{Data Explanation}
\label{sec:marker}
\hk{The CMU dataset} contains 3D positions of a set of markers, which were recorded by the mocap system at CMU. Example of a marker placement during the capture can be seen in Figure~\ref{fig:markers}. All details can be found in the dataset description \cite{CMUdata}. 

The human pose at each time-frame $t$ is represented as a vector of the marker 3D coordinates: $\mathbf{x}_t = [x_{i,t}, y_{i,t}, z_{i,t}] _{i=1:n} $, where $n$ denotes the number of markers used during the mocap collection. In the CMU data, $n=41$, and the dimensionality of a pose is $3n=123$.


A sequence of poses is denoted $\mathbf{x} = [\mathbf{x}_t]_{t=1:T}$.

\subsection{Training, Validation, and Test Data Configurations}
\label{sec:dataset_config}

\par \hk{The} \textit{validation} dataset contains 2 sequences from each of the following motions: pantomime, sports, jumping, and general motions \footnote{pantomime (subjects 32 and 54), sports (subject 86 and 127), jumping (subject 118), and general motions (subject 143)}.
\hk{The} \textit{test} dataset contains 
basketball, boxing and jump turn\footnote{102\_03 (basketball), 14\_01 (boxing), and  85\_02 (jump-turn).} sequences.  
\hk{The} \textit{training} dataset contains all the sequences not used for validation or testing,  from 25 different folders in the CMU Mocap Dataset, such as 6, 14, 32, 40, 141, 143, which include testing types as well.

Subjects from the training dataset were also present in the test and validation datasets. Generalization to the novel subjects and motion types was tested experimentally.



\pdfoutput=1
\section{Experiments}
\label{secExp}


We use the commonly used \cite{peng2015hierarchical,wang2016human,burke2016estimating} Root Mean Squared Error (RMSE) over the missing markers to measure reconstruction error. 

\textbf {Implementation Details:}
All methods were implemented using Tensorflow\cite{abadi2016tensorflow}. The code is publicly available\footnote{https://github.com/Svito-zar/NN-for-Missing-Marker-Reconstruction}.



For \textbf{training} purposes, we extract short sequences from the dataset by sliding window, then shuffle them and feed to the network. The training was done using the Adam optimizer \cite{kingma2014adam} with a batch size of 32.




The \textbf{hyperparameters} for both architectures were optimized w.r.t.~the validation dataset (Section~\ref{sec:dataset_config}) using grid search. 
Table \ref{hyperparams} contains the main hyper-parameters. 

\begin{table}
\caption{Hyperparameters for our NNs.\\ $\alpha$ is initial learning rate, $\Delta t$ is sequence length.}
\begin{tabular}
{|l|c|c|c|c|c|}
\hline
NN-type & Width & Depth & Dropout & $\alpha$ & $\Delta t$\\
\hline
LSTM& 1024 & 2 & 0.9 & 0.0002 & 64 \\
Window&  512 & 2 & 0.9 & 0.0001 & 20\\
\hline
\end{tabular}
\label{hyperparams}

\end{table}

\subsection{Comparison to the State of the Art}


First of all, the models presented above are evaluated in the same setting as most of the other random missing marker reconstruction methods. A specific amount of random markers (10\%, 20\%, or 30\%) are removed over a few time-frames and each method is applied to recover them. The length of the gap was sampled from the Gaussian distribution with mean 10 and standard deviation 5, following the state-of-the-art settings \cite{wang2016human}. The reconstruction error is measured.

There is randomness in the system; in the initialization of the network weights and choosing missing markers. Therefore, every experiment is repeated 3 times and error mean and standard deviation are measured.

Tables \ref{CompWithOthers} provide the comparison of the performance of our system with 3 state-of-the-art papers and with the simplest solution (linear interpolation) as a baseline, on 3 action classes from the CMU Mocap dataset. The experiments from \cite{burke2016estimating} were repeated by us while using the same hyperparameters as in their original paper. The results of the Wang method \cite{wang20163d} were taken from the diagram in their paper. Last, the error measures of the Peng method \cite{peng2015hierarchical} were rescaled, since in their original paper they measure it with averaging the error over all the markers, but we average only over the missing markers. 

Table \ref{CompWithOthers} shows that standard interpolation outperforms all the state-of-the-art method, including ours. A probable reason for that is that the duration of the gap is short (less than 0.1 s), so it is easy to interpolate between existing frames. 
We will, therefore, study a more challenging scenario, when markers are missing over longer periods and when more markers are missing. We can compare with the Burke method only because only they provide the implementation. 


\begin{table}[t]
\caption{Comparison to the state of the art in missing marker reconstruction. RMSE in marker position is in cm. A training set comprises all activities. $^*$The numbers from \cite{wang20163d} were extracted from a diagram.} 
\vspace{-2mm}
\centering
\subfloat[10\% of the markers in each indata frame are missing.]{
\begin{tabular}{|l|c|c|c|}
\hline
Method & Basketball & Boxing & Jump turn\\
\hline\hline
Interpolation &  0.64$\pm 0.03$ & 1.06$\pm 0.12$ & \textbf{1.74}$\pm 0.3$\\
Wang \cite{wang20163d} & \textbf{0.4}$^*$ & \textbf{0.5}$^*$ & n.a.\\
Peng \cite{peng2015hierarchical} & n.a. & n.a. & n.a.\\
Burke \cite{burke2016estimating} & 4.56 $\pm 0.17$ & 3.47$\pm 0.19$ & 15.97$\pm 1.34$ \\
Window (ours) & 2.34 $\pm 0.27$ & 2.61 $\pm 0.21$ & 4.4 $\pm 0.5$ \\
LSTM (ours) & 1.21$\pm 0.02$ &1.44$\pm 0.02$ & 2.52$ \pm 0.3$\\
\hline
\end{tabular}}
\par\medskip
\subfloat[20\% of the markers  in each indata frame are missing.]{
\begin{tabular}{|l|c|c|c|}
\hline
Method & Basketball & Boxing & Jump turn\\
\hline\hline
Interpolation & \textbf{0.67}$\pm 0.04$  & \textbf{1.09}$\pm 0.07$ & \textbf{1.91}$\pm 0.31$
 \\
Wang \cite{wang20163d} & 1.6$^*$ & 1.5$^*$ & n.a.\\
Peng \cite{peng2015hierarchical} & n.a. & 4.94 & 5.12 \\
Burke \cite{burke2016estimating} & 4.18 $\pm 0.48$ & 3.98$\pm 0.07$ & 27.1$\pm 1.21$ \\
Window (ours) & 2.42 $\pm 0.32$ & 2.77 $\pm 0.13$ & 4.3 $\pm 0.75$ \\
LSTM (ours) &  1.34$\pm 0.01$ & 1.58$\pm 0.04$ & 2.67$\pm 0.2$\\
\hline
\end{tabular}}
\par\medskip
\subfloat[30\% of the markers in each indata frame are missing.]{
\begin{tabular}{|l|c|c|c|}
\hline
Method & Basketball & Boxing & Jump turn\\
\hline\hline
Interpolation & \textbf{0.7}$\pm 0.1$ & \textbf{1.21}$\pm 0.14$ & \textbf{2.29}$\pm 0.3$
 \\
Wang \cite{wang20163d} & 0.9$^*$ & 0.9$^*$ & n.a.\\
Peng \cite{peng2015hierarchical} & n.a. & 4.36 & 4.9 \\
Burke \cite{burke2016estimating} & 4.23 $\pm 0.57$ & 4.01$\pm 0.26$ & 34.9 $\pm 2.55$ \\
Window (ours) & 2.33 $\pm 0.13$ & 2.63 $\pm 0.08$ & 4.53 $\pm 0.48$ \\
LSTM (ours) &  1.48$\pm 0.03$&1.75$\pm 0.07$ & 3.1$ \pm 0.25$\\
\hline
\end{tabular}}
\label{CompWithOthers}
\end{table}

\subsection{Gap duration analysis}

In the next experiments, we varied the length of the gap and kept the number of missing markers fixed to 5. As before we averaged the performance over 3 experiments.

\begin{figure}[t]
\centering
\includegraphics[width=0.9\linewidth]{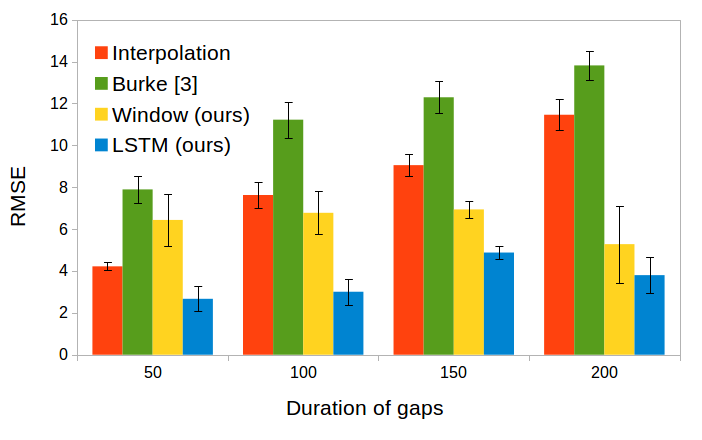}
\caption{Dependency on the duration of the gap. Basketball motion. 5 missing markers.}
\label{fig:length_analysis}
\end{figure}

Figure \ref{fig:length_analysis} shows that our methods can be applied for any length of gaps, while the performance of other methods degrades steadily with the increase of the length of the gap.  Interpolation-based methods struggle to reconstruct markers when gaps become longer. Our method, in contrast, can propagate the information about the marker position using the hidden state, hence being robust to the long gaps. 

\subsection{Very long gaps}

\begin{figure}[p]
\centering
\vspace{-2mm}
\subfloat[Basketball: 3 markers missing]{~~~\includegraphics[width=0.92\linewidth]{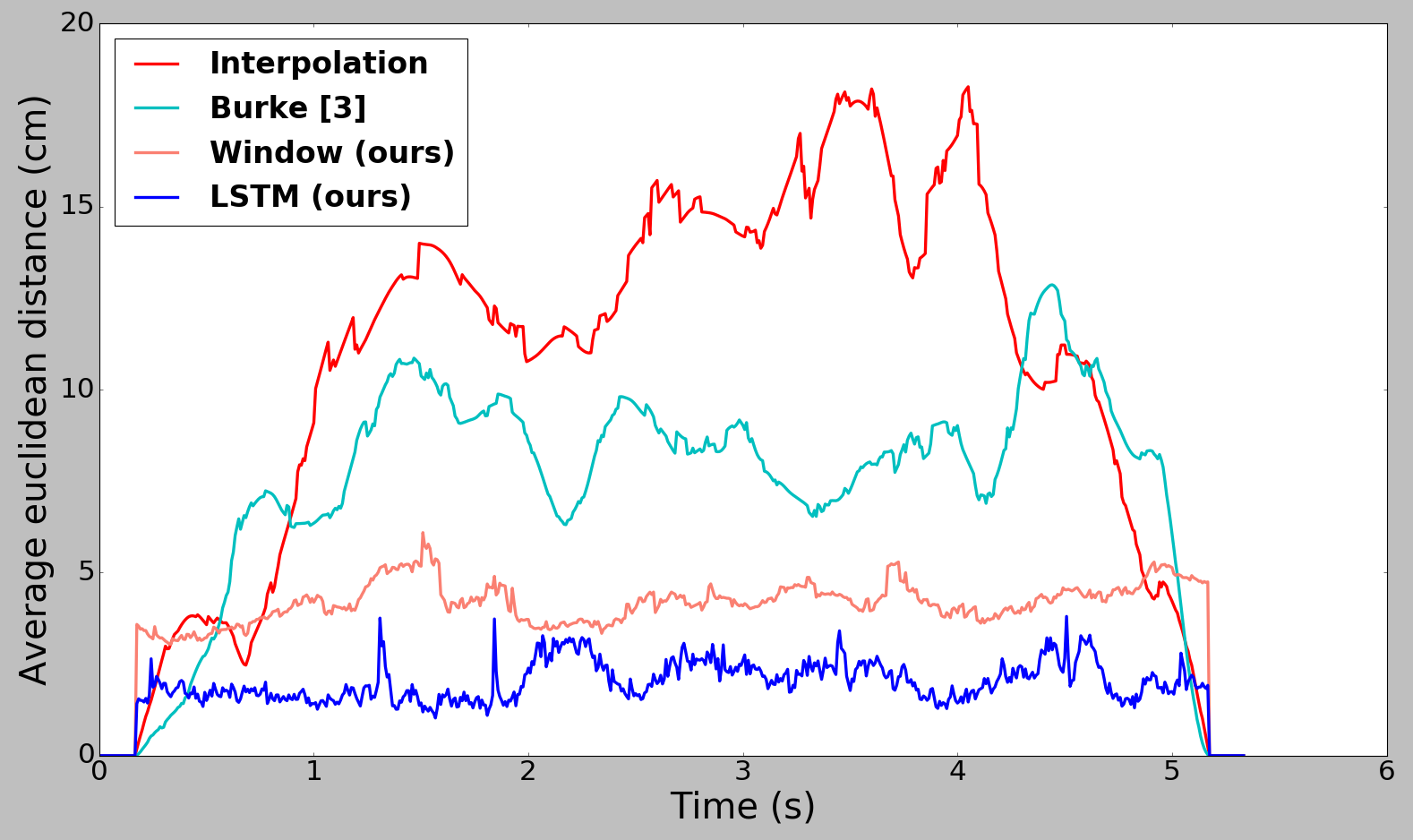}~~~}~~\\
\subfloat[Basketball: 30 markers missing]{~~~\includegraphics[width=0.92\linewidth]{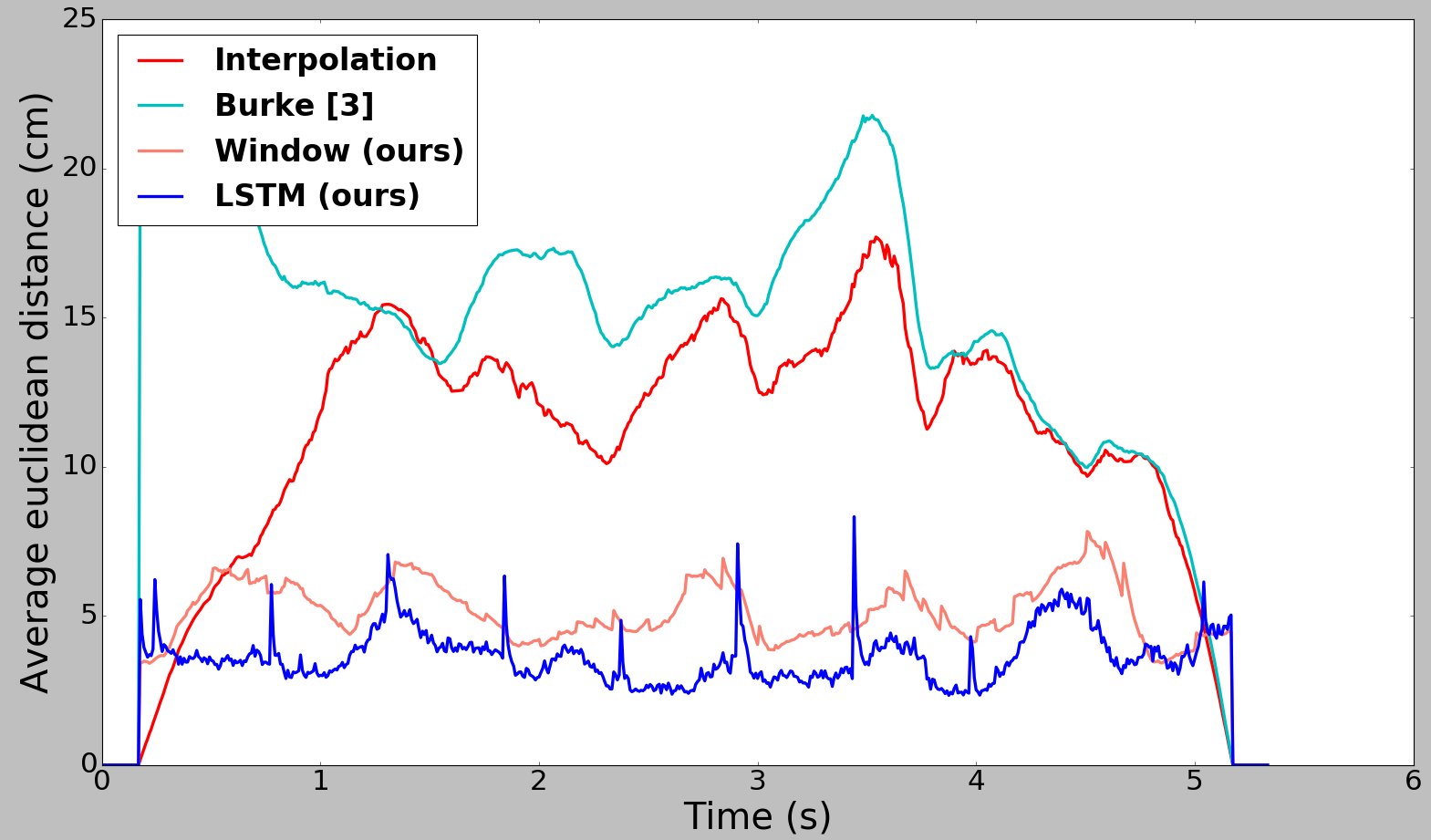}~~~}\\
\subfloat[Boxing: 3 markers missing]{~~~\includegraphics[width=0.92\linewidth]{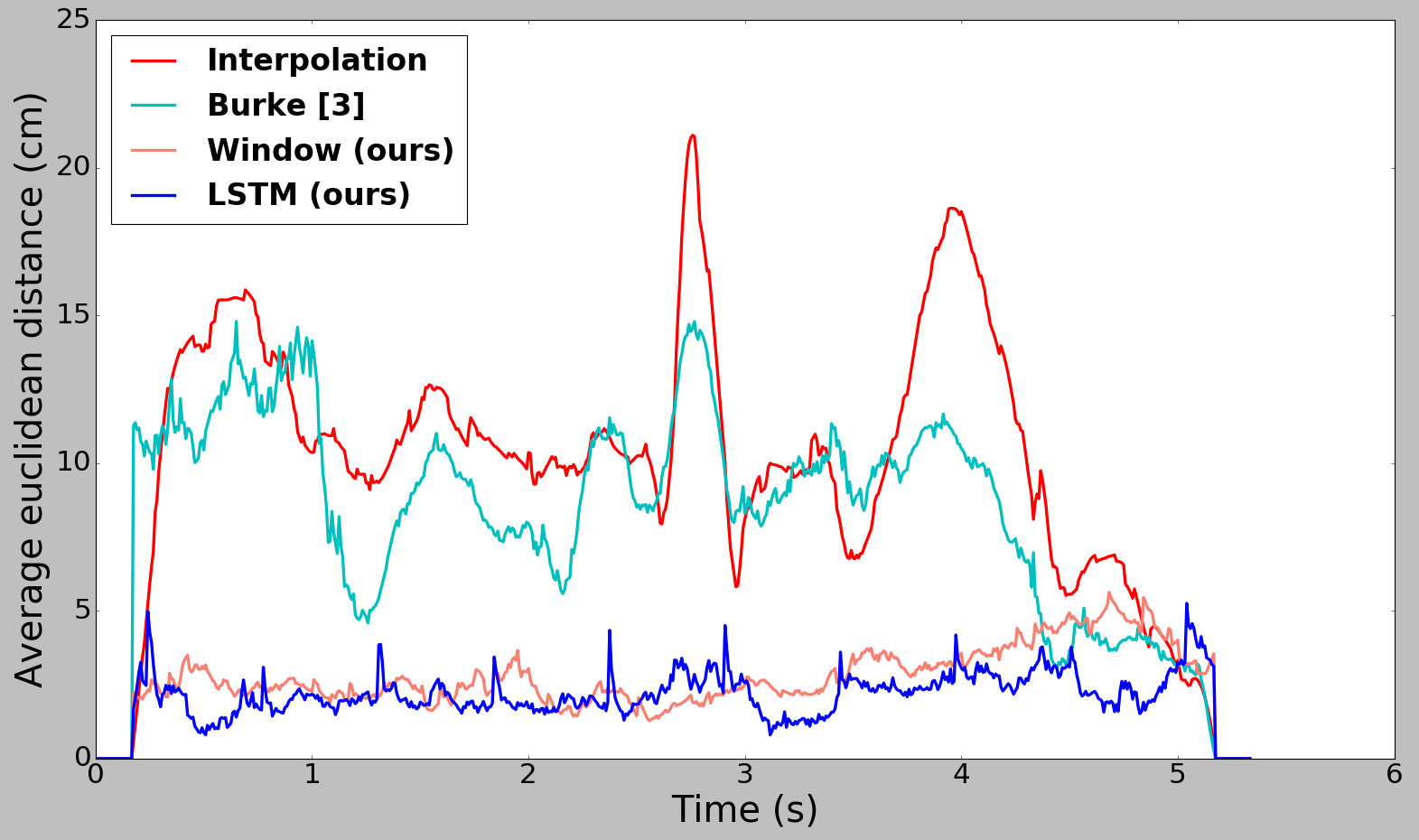}~~~}\\
\subfloat[Boxing: 30 markers missing]{~~~\includegraphics[width=0.92\linewidth]{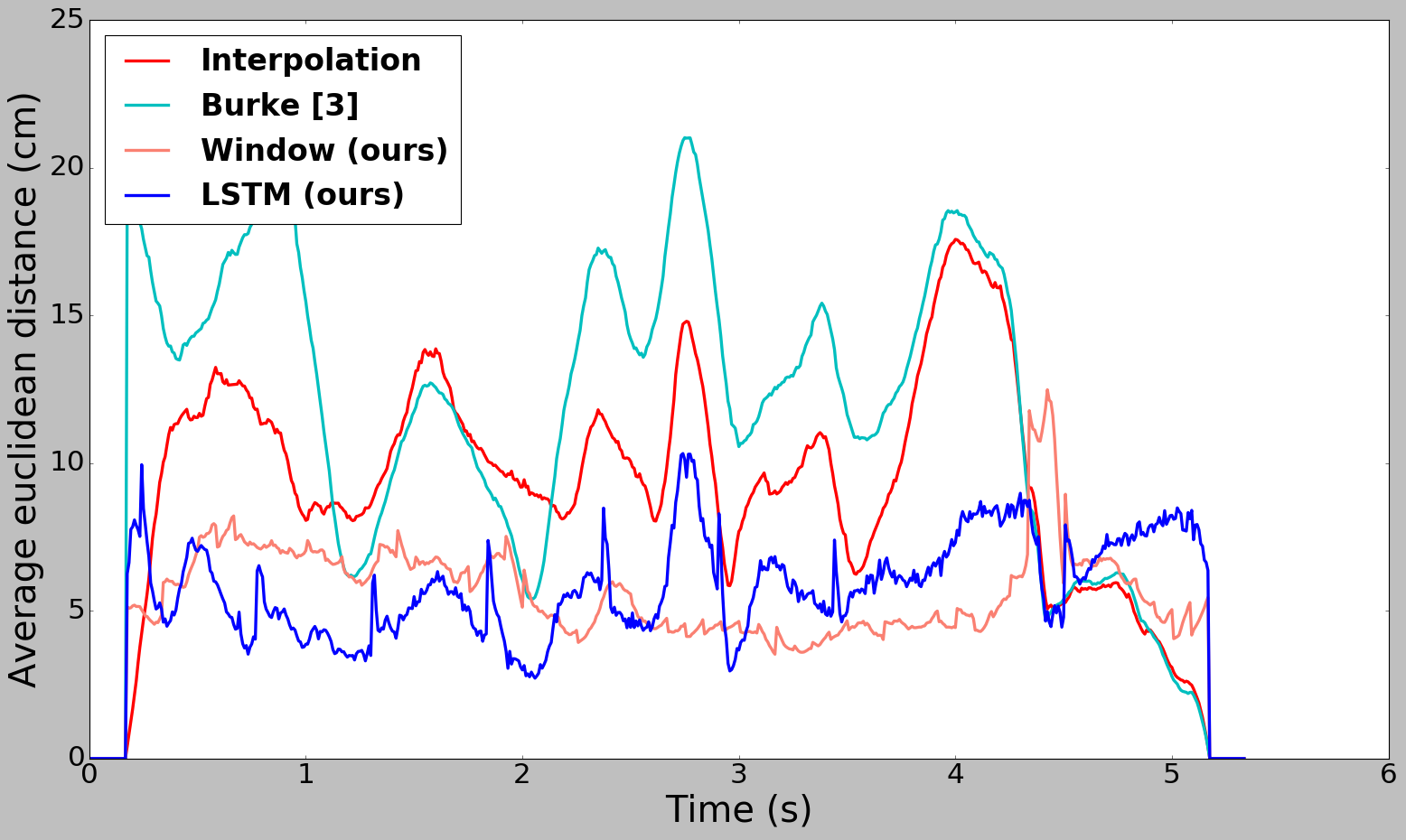}~~~}

\caption{A few markers were missing over 5 seconds. All the markers were present for 1s before and after the gap.}
\label{fig:long_gap}
\end{figure}

In the following experiment, the same markers were missing over a long period of time. The measurement period started at 1.5 seconds into the clip to avoid artifacts. Then for 1 second, all markers were present, followed by a 5 second window where certain markers were missing for the entire time. Afterwards, all the markers were present again. Each experiment was repeated 5 times and the mean result was registered.

We can clearly see in Figure \ref{fig:long_gap} that while interpolation and Burke\cite{burke2016estimating} are quickly losing track of the markers, our methods stay stable and accurate. This hold for all the scenarios. Figure \ref{fig:long_gap}(b,d) illustrates that all methods except interpolation are degrading significantly when most of the markers are missing. That indicates that those methods are using information about the other markers, not only about the past or the future of a particular marker.

\subsection{Visualization of the Results}

\begin{figure}[t]
\centering
\vspace{-2mm}
\subfloat[Ground truth markers]{~~~\includegraphics[width=0.815\linewidth]{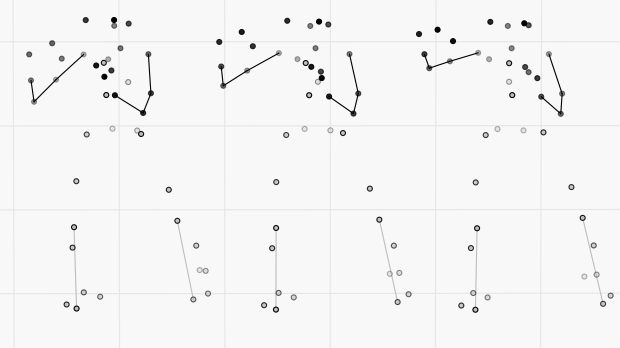}~~~}~~\\
\subfloat[15 (our of 41) markers are missing]{~~~\includegraphics[width=0.815\linewidth]{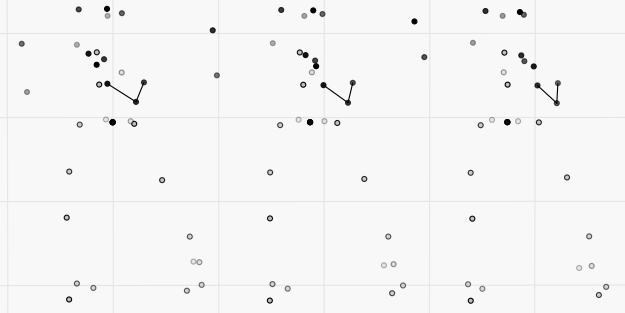}~~~}\\
\subfloat[Burke{[3]} reconstruction result]{~~~\includegraphics[width=0.815\linewidth]{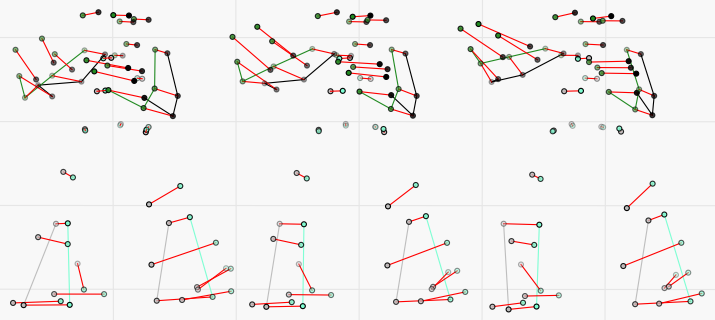}~~~}\\
\subfloat[LSTM (ours) reconstruction result]{~~~\includegraphics[width=0.815\linewidth]{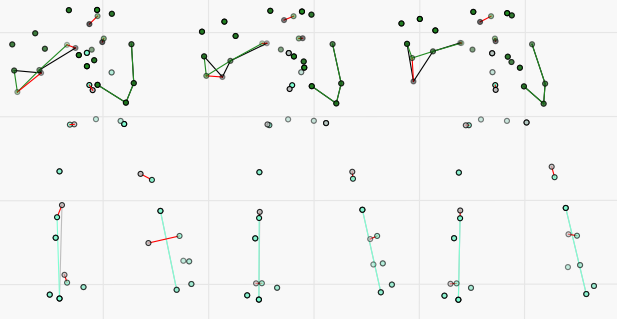}~~~}
\caption{Three keyframes from the boxing test sequence, illustration of the reconstruction using the Burke and LSTM (ours) methods.}

\label{fig:VisualResults}
\end{figure}

Figure~\ref{fig:VisualResults} illustrates the reconstruction results for one of the test sequences, boxing. 
The subject is boxing with their right arm. The observed marker cloud (Figure~\ref{fig: VisualResults}b) misses 15 markers. 
Our reconstruction result (Figure~\ref{fig: VisualResults}d) is visually close to the ground truth (Figure~\ref{fig: VisualResults}a), which is also supported by the numerical errors.

\subsection{Generalization}

\begin{table}
\caption{Generalization test for the LSTM network.  20\% of markers missing. Complete dataset contains motions from all the subjects and from all types. Then all motion with the same subject as in testing were removed. Finally all the recordings with the same type of motion very removed. Reconstruction error in cm is measured.}
\vspace{-3mm}
\begin{tabular}{|l|c|c|}
\hline
Motion / dataset & Basketball & Boxing \\
\hline
Complete & 7.9 $\pm{ 0.14}$ & 2.08 $\pm{ 0.5}$  \\
w/o the subject &  9.93 $\pm{ 0.96}$ & 2.13  $\pm{ 0.42}$  \\
w/o the motion &  8.54  $\pm{ 1.04}$  & 2.57  $\pm{ 1.18}$  \\
\hline
\end{tabular}
\label{generalization_lstm}

\end{table}

\begin{table}
\caption{Generalization test for the Window-based network. 20\% of markers missing. The same setup as in Table \ref{generalization_lstm}.} 
\vspace{-3mm}
\begin{tabular}{|l|c|c|}
\hline
Motion / dataset & Basketball & Boxing \\
\hline
Complete & 5.59 $\pm{ 0.29}$ & 3.54 $\pm{ 0.15}$  \\
w/o the subject &  5.68 $\pm{ 0.48}$ & 4.37  $\pm{ 1.13}$  \\
w/o the motion &  6.52  $\pm{ 0.54}$  & 4.58  $\pm{ 1.51}$  \\
\hline
\end{tabular}
\label{generalization_window}

\end{table}

Up to now, our models, as well as the baselines, have been trained with all motions and all individuals.

In the final experiments, we evaluated the generalization capability with respect to motion type and individual. To this end, we removed all the recordings of the test subject from the training data.
 Furthermore, for each test motion, we created a training set where this motion was removed. 
We then evaluated our networks while having 20\% of the markers missing for gaps of 100 frames (almost 1 second).

Table \ref{generalization_lstm} illustrates the results for the LSTM-based network and Table \ref{generalization_window} for the Window-based method. We can observe that the performance drop is not dramatic: it is less than 25\% and depends on the motion type and the network architecture. It is important to note that the variance is significantly higher for the "generalization" scenarios, meaning that the system is less stable.

This experiment indicates that our systems can recover unseen motions, performed by unseen individuals, albeit 
with slightly worse performance.


\pdfoutput=1
\section{Discussion and CONCLUSION}
\label{Sec:Disc}

The experiments presented above show that the proposed methods can compensate for missing markers better than the state of the art, when the gap is long, especially when the motion is complex. 

Our method is not relying on future frames, unlike most of the alternatives. That property makes it suitable for on-line usages when the markers are being reconstructed as they are collected.

Another notable property of the proposed method is that it can recover markers which are missing over a long period of time. 

LSTM-based architecture is modeling the correlation is the human body to recover missing markers better than a window-based architecture, accordingly to our experiments.

In summary, the proposed methods can be used to recover markers over many frames in an accurate and stable way.




\section{Acknowledgements}
Authors would like to thank Simon Alexanderson and Judith Butepage for the useful discussions. This PhD project is supported by Swedish Foundation for Strategic Research  Grant No.: RIT15-0107. 
The data used in this project was obtained from mocap.cs.cmu.edu.
The database was created with funding from NSF EIA-0196217.

\bibliographystyle{ACM-Reference-Format}
\bibliography{sample-bibliography}

\end{document}